\documentclass[conference]{IEEEtran}
\usepackage[top=0.7in, bottom=0.96in, left=0.65in, right=0.65in]{geometry}
\IEEEoverridecommandlockouts

\usepackage{cite}
\usepackage{amsmath,amssymb,amsfonts}
\usepackage{algorithmic}
\usepackage{graphicx}
\usepackage{textcomp}
\usepackage{xcolor}
\usepackage{algorithm}
\usepackage{algorithmic}
\usepackage{booktabs}
\usepackage{multirow}
\usepackage{graphicx}
\usepackage{subcaption}
\usepackage{caption}
\usepackage{balance}

\usepackage[font=small,labelfont=bf]{caption}
\usepackage[font=footnotesize]{subcaption}

\usepackage{amsthm}

\theoremstyle{plain}

\theoremstyle{definition}

\theoremstyle{remark}

\def\BibTeX{{\rm B\kern-.05em{\sc i\kern-.025em b}\kern-.08em
    T\kern-.1667em\lower.7ex\hbox{E}\kern-.125emX}}
\begin{document}

\title{LWM-Temporal: Sparse Spatio-Temporal Attention for Wireless Channel Representation Learning}

\author{
Sadjad Alikhani$^{1}$, Akshay Malhotra$^{2}$, Shahab Hamidi-Rad$^{2}$, Ahmed Alkhateeb$^{1}$\\
$^{1}$\{alikhani, alkhateeb\}@asu.edu, $^{2}$\{akshay.malhotra, shahab.hamidi-rad\}@interdigital.com
}

\maketitle

\begin{abstract}
LWM-Temporal is a new member of the Large Wireless Models (LWM) family that targets the \emph{spatiotemporal} nature of wireless channels. Designed as a task-agnostic foundation model, LWM-Temporal learns universal channel embeddings that capture mobility-induced evolution and are reusable across various downstream tasks. To achieve this objective, LWM-Temporal operates in the angle-delay-time domain and introduces Sparse Spatio-Temporal Attention (SSTA), a propagation-aligned attention mechanism that restricts interactions to physically plausible neighborhoods, reducing attention complexity by an order of magnitude while preserving geometry-consistent dependencies. LWM-Temporal is pretrained in a self-supervised manner using a physics-informed masking curriculum that emulates realistic occlusions, pilot sparsity, and measurement impairments. Experimental results on channel prediction across multiple mobility regimes show consistent improvements over strong baselines, particularly under long horizons and limited fine-tuning data, highlighting the importance of geometry-aware architectures and geometry-consistent pretraining for learning transferable spatiotemporal wireless representations.
\end{abstract}


\section{Introduction}

Next-generation wireless systems exploit large antenna arrays, mid-band to sub-THz spectra, and dense deployments to meet stringent throughput and latency targets~\cite{8808168}. These gains, however, hinge on accurately modeling a rapidly time-varying channel shaped by propagation geometry and mobility: as users move, multipath components drift in angle and delay, undergo birth/death under occlusions, and induce correlated Doppler shifts~\cite{Tse_Viswanath_2005}. Capturing this high-dimensional spatiotemporal evolution is critical for massive MIMO, mmWave beam tracking, and integrated sensing and communication~\cite{6457363}. This motivates transferable representation learning that encodes propagation-consistent dynamics, i.e., the constrained angle-delay-Doppler evolution induced by a latent physical scene, in a domain where such structure is explicit.

\textbf{Prior Work:} Conventional channel modeling and simulation relies on stochastic models (e.g., tapped-delay-line/CDL and 3GPP UMa/UMi)~\cite{3gppTR38901} which reproduce aggregate statistics and Doppler-driven phase evolution but abstract away explicit geometry~\cite{morais2025comparingstochasticraytracingdatasets}. As a result, they miss trajectory-consistent angle-delay dynamics (cluster motion and path birth/death), and learning systems trained on them often capture only Doppler-level statistics, limiting long-horizon forecasting~\cite{morais2025comparingstochasticraytracingdatasets}.
Learning-based methods, including recurrent and hybrid architectures, have achieved strong performance in channel estimation and short-term prediction~\cite{8813020,POTTER2010440,chen2021temporal}. However, they are typically trained in task- and scenario-specific regimes, with limited ability to transfer across environments or mobility conditions; moreover, as the prediction horizon grows, error accumulation and limited effective memory hinder capturing long-range dependencies and abrupt structural changes such as blockage-induced path birth/death.
Recent Transformer-based foundation models learn transferable embeddings via large-scale self-supervision~\cite{alikhani2024large,11140266,wifo2024,llm4cp2024}. Yet scaling to long sequences remains problematic: dense self-attention is quadratic in length, while generic efficiency mechanisms inherited from NLP and vision/video~\cite{tay2020efficient,child2019generating,beltagy2020longformer,zaheer2020big,kitaev2020reformer,roy2021efficient,wang2020linformer,choromanski2021rethinking,liu2021swin,bertasius2021is} can discard propagation-structured long-range interactions essential for geometry-aware prediction. This creates an unfavorable trade-off between expressiveness and computational tractability, motivating propagation-aligned efficiency designs.

\begin{figure*}[t]
\centering
\includegraphics[width=\textwidth]{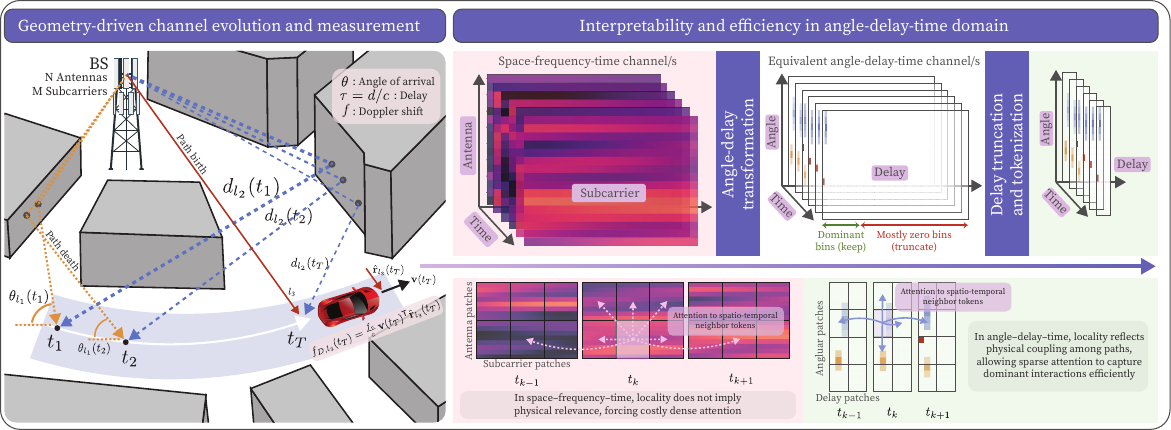}
\caption{This figure illustrates the geometry-driven evolution of wireless channels in dynamic environments and its role in efficient modeling. As users move, propagation paths evolve according to the underlying geometry, inducing structured changes in angle, delay, and Doppler. Transforming channels into the angle-delay-time domain exposes these physical and temporal structures in a more interpretable and sparser representation, enabling efficient tokenization and sparse attention over physically plausible interactions that are less explicit in the original space-frequency-time domain.}
\label{fig:lwm_temporal_overview}
\end{figure*}

\textbf{Contribution:} We present LWM-Temporal, a new member of the Large Wireless Models (LWMs) family that extends the original LWM from single-snapshot modeling to spatiotemporal channel sequences. Relative to prior LWM, LWM-Temporal (i) models channel trajectories, (ii) introduces Sparse Spatio-Temporal Attention (SSTA) for near-linear scaling over long horizons, (iii) adopts an angle-delay-time tokenization that exposes propagation primitives and their evolution, and (iv) uses a physics-informed masking curriculum tailored to occlusions and sparse sensing. The model is pretrained on large-scale, geographically diverse ray-traced trajectories to learn geometry-consistent dynamics. Our contributions are:
\begin{enumerate}
    \item A physics-informed sparse attention mechanism (SSTA) that restricts interactions to local and temporally correlated tokens in the angle-delay-time domain, achieving near-linear complexity.
    \item A self-supervised pretraining framework using realistic masking patterns on ray-traced data, enabling robust learning of joint angle-delay-Doppler evolution.
    \item State-of-the-art channel prediction performance across mobility regimes, demonstrating strong zero-shot generalization and highlighting the necessity of geometry-aware architectures paired with geometry-consistent data.
\end{enumerate}
LWM-Temporal provides an efficient, physically grounded foundation for transferable spatiotemporal channel representation learning. \footnote{Pretrained model, training scripts, tutorials, demo, and the dynamic scene and data generation pipeline are available at \texttt{https://huggingface.co/wi-lab}.}

\section{Problem Formulation}

Next-generation wireless communication and sensing systems require models that can learn transferable representations of channel state information (CSI), capturing complex dependencies across space, frequency, and time while respecting the physics of wireless propagation. Let
$
\mathbf{H}_t \in \mathbb{C}^{N \times M}
$
denote the complex-valued channel matrix observed at time index $t$, where $N$ is the number of base-station antennas and $M$ is the number of subcarriers. Depending on the application and measurement availability, the model input may consist of a single channel realization or a sequence of channel observations as follows
\begin{equation}
\mathcal{H}_{t_0:T} = \{\mathbf{H}_{t_0}, \mathbf{H}_{t_0+1}, \ldots, \mathbf{H}_{T}\},
\end{equation}
where the sequence length $T - t_0 + 1$ is variable.

The objective of this work is to learn a task-agnostic representation function
\begin{equation}
f_\theta:\ \mathcal{H}_{t_0:T} \longrightarrow \mathbf{Z} \in \mathbb{R}^{L \times D},
\end{equation}
where $\mathbf{Z}$ is a latent token sequence that encodes the salient spatial, spectral, and temporal structure of the input channel sequence. Here, $L$ denotes the number of tokens in $\mathbf{Z}$ and $D$ is the embedding dimension. The representation $\mathbf{Z}$ is designed to transfer across tasks via a lightweight head
\begin{equation}
\hat{\mathbf{y}} = g_\phi\!\big(\mathcal{S}(\mathbf{Z})\big),
\end{equation}
where $\mathcal{S}(\cdot)$ aggregates the representation into a fixed-dimensional summary and $g_\phi(\cdot)$ is task-specific. This formulation covers: \textbf{reconstruction} (recover missing/noisy/compressed CSI by minimizing $\mathcal{L}_{\text{rec}}=\|\mathbf{H}-\hat{\mathbf{H}}\|^2$), \textbf{forecasting} (predict future channels $\hat{\mathbf{H}}_{T+1:T+\Delta}=g_\phi(\mathcal{S}(\mathbf{Z}))$), \textbf{discriminative/semantic} inference (classification/regression/clustering from CSI), and \textbf{decision-oriented} control (e.g., beam selection or resource allocation) without explicit reconstruction.

The challenge is to design $f_\theta$ so that $\mathbf{Z}$ captures intrinsic spatiotemporal structure while remaining efficient and robust to variable-length inputs and deployment shifts.

\section{Method} \label{sec:method}

LWM-Temporal extends masked channel modeling~\cite{alikhani2024large} with inductive biases aligned with wireless propagation geometry. It aligns the representation domain, attention, and pretraining with the geometry-driven evolution of paths by operating in the angle-delay-time (AD-$t$) domain and using three stages: transformation and tokenization, sparse spatio-temporal attention over physically plausible interactions, and physics-informed self-supervised pretraining. These components, shown in Fig.~\ref{fig:lwm-temporal}, yield efficient, transferable spatiotemporal representations for diverse downstream tasks.

\subsection{Angle-Delay-Time Domain Transformation}
\label{subsec:adt_structure}

Learning in the space-frequency-time domain is challenging because proximity between coefficients is only weakly tied to propagation, and modeling interactions across antennas, subcarriers, and time scales scales poorly with sequence length. We therefore transform the channel sequence into the angle-delay-time domain
\begin{equation}
\mathbf{H}(t,\theta,\tau)
=
\mathcal{F}_{\mathrm{del}}^{-1}\!\left\{
\mathcal{F}_{\mathrm{ang}}\!\left[\mathbf{H}(t,n,m)\right]
\right\},
\end{equation}
where $\mathcal{F}_{\mathrm{ang}}$ is an $N$-point DFT over the antenna index $n$ that maps to beamspace (angle) bins, and $\mathcal{F}_{\mathrm{del}}^{-1}$ is an $M$-point IDFT over the subcarrier index $m$ that maps to delay bins~\cite{wen2018deeplearningmassivemimo}.
In AD-$t$, channel energy is sparse and interpretable, and temporal evolution is structured: dominant components drift smoothly in $(\theta,\tau)$ with occasional birth/death under occlusions. These properties motivate geometry-aligned representation learning that restricts computation to physically plausible interactions. LWM-Temporal instantiates this principle via sparse spatio-temporal attention that follows AD-$t$ evolution efficiently.

\subsection{Tokenization}
Each time step $t$ yields an angle-delay frame $\mathbf{H}(t,\theta,\tau)\in\mathbb{C}^{N\times M}$ obtained from the space-frequency channel, with angular and delay grids matching the antenna and subcarrier resolutions. Since most energy concentrates in the early delay bins, we truncate to the first $W$ taps and retain $\mathbf{H}(t)\in\mathbb{C}^{H\times W}$ with $H=N$ and $W\le M$~\cite{wen2018deeplearningmassivemimo}, reducing input size with minimal performance loss. Each truncated frame is partitioned into $(P_h\times P_w)$ patches, flattened, and projected to $\mathbb{R}^{D}$. With temporal length $T$, the number of tokens per frame and per sequence are
\begin{align}
N_{\text{tok}} &= \frac{H}{P_h}\cdot\frac{W}{P_w}, &
S &= T\cdot N_{\text{tok}} + 1 ,
\end{align}
where the additional $+1$ corresponds to a global \texttt{[CLS]} token appended to summarize the entire spatiotemporal window. We normalize token channels to stabilize training under high dynamic range and sparse observations.

\subsection{Rotary Positional Encoding (RoPE)}
To encode relative positions along the flattened token sequence, we apply RoPE to per-head queries and keys.
For token $i$ with sequence index $p_i$ and head dimension $d$, define $\mathbf{q}_i=\mathbf{W}_q\mathbf{x}_i\in\mathbb{R}^d$ and $\mathbf{k}_i=\mathbf{W}_k\mathbf{x}_i\in\mathbb{R}^d$.
RoPE applies a position-dependent rotation
\begin{align}
\mathbf{q}_i^{(\mathrm{rot})} &= \mathbf{R}(p_i)\,\mathbf{q}_i, &
\mathbf{k}_i^{(\mathrm{rot})} &= \mathbf{R}(p_i)\,\mathbf{k}_i,
\end{align}
where $\mathbf{R}(p_i)$ is block-diagonal with $2\times 2$ rotation blocks at logarithmically spaced frequencies.
This construction preserves the relative-position property such that
$\langle \mathbf{q}_i^{(\mathrm{rot})}, \mathbf{k}_j^{(\mathrm{rot})} \rangle$
depends on the offset $\Delta p_{ij}=p_i-p_j$, providing a simple relative encoding that complements SSTA by biasing attention toward smooth token-to-token drifts.

\subsection{Sparse Spatio-Temporal Attention (SSTA)}

\begin{figure}[t]
    \centering
    \includegraphics[width=\columnwidth]{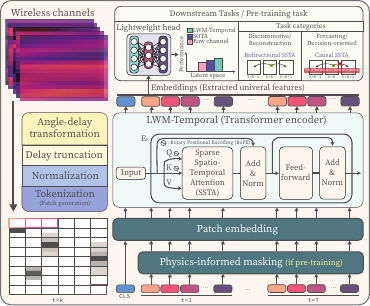}
   \caption{\textbf{LWM-Temporal overview.}
Channels are transformed to angle-delay, tokenized, and pretrained with physics-informed masking and RoPE-based sparse spatiotemporal attention (SSTA) to produce embeddings. SSTA is bidirectional for reconstruction and causal for forecasting to prevent leakage.}
    \label{fig:lwm-temporal}
\end{figure}

\begin{figure}[t]
    \centering
    \begin{subfigure}[t]{0.32\columnwidth}
        \centering
        \includegraphics[width=\linewidth]{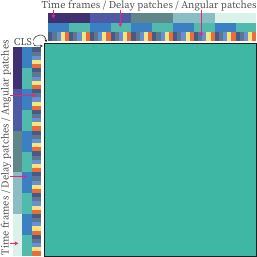}
        \caption{Full Attention}
        \label{fig:sub1}
    \end{subfigure}
    \hfill
    \begin{subfigure}[t]{0.32\columnwidth}
        \centering
        \includegraphics[width=\linewidth]{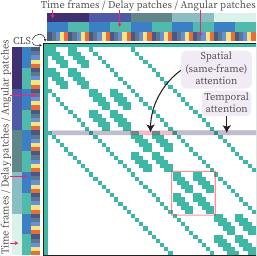}
        \caption{SSTA w/o Routing}
        \label{fig:sub2}
    \end{subfigure}
    \hfill
    \begin{subfigure}[t]{0.32\columnwidth}
        \centering
        \includegraphics[width=\linewidth]{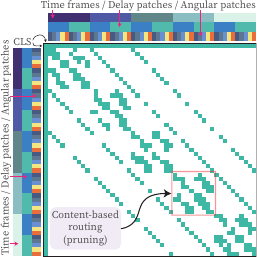}
        \caption{SSTA}
        \label{fig:sub3}
    \end{subfigure}
    \caption{\textbf{Attention patterns for angle-delay sequences.}
Colored strips show the flattened token order (time, angle, delay); each matrix shows allowable attention links.
(a) Full attention: all-to-all, \(\mathcal{O}(N^2)\).
(b) SSTA: predefined local neighbors, \(\mathcal{O}(KN)\).
(c) SSTA (routed): \(K_r\) adaptive neighbors, \(\mathcal{O}(K_rN)\).}
    \label{fig:att_maps}
\end{figure}

Dense attention over $S$ tokens is quadratic and agnostic to propagation physics. AD-$t$ channels are \emph{sparse and structured}: energy concentrates in a few blobs (clusters), which move smoothly over time and couple angle, delay, and motion. As shown in Fig. \ref{fig:att_maps}, SSTA encodes this by restricting each query to a \emph{structured neighborhood} that (i) captures local AD coupling within a frame and (ii) follows feasible temporal trajectories (“corridors”) across frames. This yields near-linear complexity while preserving the interactions that matter physically.

\textbf{Neighborhood definition:}
For a token at $(t,h,w)$,
% \begin{subequations}
\begin{align}
&\mathcal{N}(t,h,w) \;=\;
\underbrace{\mathcal{N}_{\text{local}}(t,h,w)}_{\text{same-frame window}}
\;\cup\;
\underbrace{\mathcal{N}_{\text{temp}}(t,h,w)}_{\text{temporal corridors}}, \\
&\mathcal{N}_{\text{local}}(t,h,w) = \{(t,h',w'):\ |h'{-}h|\!\le r_h,\ |w'{-}w|\!\le r_w\}, \nonumber \\
&\mathcal{N}_{\text{temp}}(t,h,w)  = \{(t{+}\Delta t,h',w'):\ \Delta t\!\in\!\mathcal{T},\nonumber\\
& \hspace{2.5cm}|h'{-}h|\!\le\! \gamma_h|\Delta t|,\ |w'{-}w|\!\le\! \gamma_w|\Delta t|\} \nonumber ,
\end{align}
% \end{subequations}
where $r_h,r_w$ set same-frame radii; $\mathcal{T}$ is a small set of temporal offsets; and $\gamma_h,\gamma_w$ bound drift per frame (AD motion constraints). This encodes feasible blob motion in AD-$t$.
The temporal offset set $\mathcal{T}$ controls whether tokens attend to the past only or to both past and future frames. 
For bidirectional representation learning, we use a symmetric set $\mathcal{T}\subseteq\{\pm 1,\pm 2,\ldots\}$, enabling each token to aggregate context from both directions in time. 
For forecasting or other generative settings, we enforce causality by restricting $\mathcal{T}\subseteq\{-1,-2,\ldots\}$, so tokens attend only to earlier frames. 
In our framework, pretraining uses bidirectional attention, while downstream generative tasks are fine-tuned with causal attention.

\textbf{Attention update and complexity:}
Let $\mathbf{x}_i$ be token $i$, $\mathcal{N}(i)$ its neighbor set, and $d$ the per-head dimension.
Using the RoPE-applied projections above, for $j\in\mathcal{N}(i)$ we compute
\begin{equation}
e_{ij}=\frac{\big(\mathbf{q}_i^{(\mathrm{rot})}\big)^{\top}\mathbf{k}_j^{(\mathrm{rot})}}{\sqrt{d}},
\qquad
\alpha_{ij}=\frac{\exp(e_{ij})}{\sum_{j'\in\mathcal{N}(i)} \exp(e_{ij'})}.
\end{equation}
Given the value projection $\mathbf{v}_j=\mathbf{W}_v\mathbf{x}_j$, the output token is
\begin{equation}
\hat{\mathbf{x}}_i=\sum_{j\in\mathcal{N}(i)} \alpha_{ij}\,\mathbf{v}_j .
\end{equation}
If $K=|\mathcal{N}(i)|\ll S$, SSTA costs $\mathcal{O}(B S K D)$ versus $\mathcal{O}(B S^{2} D)$ for dense attention, where $B$ is batch size, $S$ is sequence length, and $D$ is the embedding dimension.

\textbf{Top-$K_r$ routing:} We sparsify attention within $\mathcal{N}(i)$ by retaining only the $K_r$ neighbors with the largest logits $e_{ij}$. Specifically,
\begin{equation}
\mathcal{N}_{K_r}(i)=\mathrm{TopK}_{j\in\mathcal{N}(i)}\!\left(e_{ij}\right),
\end{equation}
and we renormalize $\alpha_{ij}$ over $\mathcal{N}_{K_r}(i)$. The routed neighborhood size is chosen as
$
K_r=\mathrm{clip}\!\left(\left\lfloor f\,|\mathcal{N}(i)| \right\rfloor,\ K_{\min},\ K_{\max}\right),
$
where $f\in(0,1]$ is a routing ratio that specifies the fraction of neighbors retained, and $K_{\min}$ and $K_{\max}$ bound the minimum and maximum number of selected neighbors. This sharpens attention and reduces the effective per-token neighborhood size from $|\mathcal{N}(i)|$ to $K_r$.

\subsection{Backbone Architecture}
We stack $E$ pre-norm residual blocks, as shown in Fig.~\ref{fig:lwm-temporal}. Given token embeddings $\mathbf{Z}\in\mathbb{R}^{S\times D}$, each block applies
\begin{subequations}
\begin{align}
\mathbf{Z} &\leftarrow \mathbf{Z} + \mathrm{SSTA}(\mathrm{LN}(\mathbf{Z})),\\
\mathbf{Z} &\leftarrow \mathbf{Z} + \mathrm{MLP}(\mathrm{LN}(\mathbf{Z})),
\end{align}
\end{subequations}
where the MLP uses an expansion ratio $r$ with SwiGLU activations. A lightweight linear reconstruction head predicts masked tokens (and unpatches them to the angle-delay grid when needed), while downstream heads (e.g., prediction, estimation, and feedback) reuse the same pretrained backbone.

By (i) encoding AD-$t$ geometry and relative positions (RoPE) and (ii) restricting attention to physics-plausible neighborhoods with saliency-aware pruning (SSTA), LWM-Temporal achieves near-linear attention cost while preserving the long-range, physically meaningful dependencies that drive wireless channel behavior.

\begin{table}
\centering
\caption{Pretraining hyperparameters}
\label{tab:pretrain-hparams}
\setlength{\tabcolsep}{4pt}
\footnotesize
\begin{tabular}{ll@{\hspace{8pt}}ll}
\toprule
\bf Hyperparameter & \bf Value & \bf Hyperparameter & \bf Value\\
\midrule
Depth $E$ & 12 & Heads $H$ & 8 \\
Embed dim $D$ & 32 & MLP ratio $r$ & 4 \\
Patch $(P_h,P_w)$ & $(1,1)$ & Activation & SwiGLU \\
Loss & NMSE & Mask ratio $\rho$ & $0.60$ (curr.) \\
Same-frame window & $3\times 3$ & Temporal offsets & $\pm\{1,2,3,4\}$ \\
Top-$K$ (max) & 64 & Routing fraction $f$ & 0.2 \\
Optimizer & AdamW & Grad clip & 1.0 \\
LR schedule & Cosine & Warmup ratio & 0.1 \\
Normalization & per-sample RMS & Positional enc. & RoPE \\
\bottomrule
\end{tabular}
\end{table}

\section{Pretraining}

LWM-Temporal is pretrained to be robust to sparse, corrupted, and imperfect channel observations. Each input sequence is RMS-normalized per sample to stabilize dynamic range. We then apply lightweight stochastic augmentations, including random phase rotation, amplitude scaling, AWGN injection, and masking, to emulate practical impairments (noise, fading variations, intermittent sensing, and partial sensor/hardware dropouts). The model is trained to reconstruct masked tokens by minimizing an NMSE objective (defined below) using AdamW with a cosine learning-rate schedule, mixed precision, and multi-GPU training.

\subsection{Physics-Informed Masked Channel Modeling (PI-MCM)}
\label{sec:physics_informed_masking}

\begin{figure}[t]
    \centering
    \begin{subfigure}[t]{0.15\columnwidth}
        \centering
        \includegraphics[width=\linewidth, height=2.5cm]{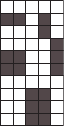}
        \caption{Rect}
        \label{fig:rectangular_mask}
    \end{subfigure}
    \hfill
    \begin{subfigure}[t]{0.15\columnwidth}
        \centering
        \includegraphics[width=\linewidth, height=2.5cm]{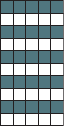}
        \caption{Comb}
        \label{fig:comb_mask}
    \end{subfigure}
    \hfill
    \begin{subfigure}[t]{0.15\columnwidth}
        \centering
        \includegraphics[width=\linewidth, height=2.5cm]{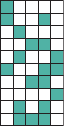}
        \caption{Rand}
        \label{fig:random_mask}
    \end{subfigure}
    \hfill
    \begin{subfigure}[t]{0.48\columnwidth}
        \centering
        \includegraphics[width=\linewidth, height=2.5cm]{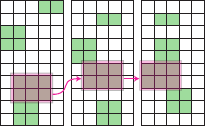}
        \caption{Spatiotemporal tube}
        \label{fig:tube_mask}
    \end{subfigure}

    \caption{Visualization of four physics-informed masking strategies. 
    The first three masks (rectangular, pilot-lattice, random) operate within individual frames,
    while the spatiotemporal tube mask creates a continuous occluded region that evolves across multiple time steps.}
    \label{fig:masking_patterns}
\end{figure}

We design masking patterns that resemble realistic sparsity and occlusions in the angle-delay-time token grid, as shown in Fig. \ref{fig:masking_patterns}. Let $(t,h,w)\in\{1,\ldots,T\}\times\{1,\ldots,H\}\times\{1,\ldots,W\}$ index tokens after patching. We define a binary mask $\mathbf{M}\in\{0,1\}^{T\times H\times W}$ where $\mathbf{M}(t,h,w)=1$ masks a token and $\mathbf{M}(t,h,w)=0$ keeps it visible. Flattening yields $L=THW$ tokens; for a target mask ratio $\rho$, we mask exactly $K=\lfloor \rho L\rfloor$ tokens by trimming/padding uniformly at random so that all masking modes use the same supervision budget. Per batch, we sample a masking mode $m\in\{\text{rect},\text{tube},\text{comb},\text{random}\}$ using an \emph{auto} mixture (with a small random branch for diversity), and apply the \emph{same} mask to both complex channels (e.g., $\Re,\Im$) to preserve physical consistency.

\textbf{Rectangular (frame-local):}
We mask a compact block within selected frames: $\mathbf{M}(t^\star,h,w)=1$ for $h\in[h_0,h_0{+}k_h)$ and $w\in[w_0,w_0{+}k_w)$. We bias the block shape to be anisotropic (typically narrower in delay) to resemble localized cluster occlusions or partial dropouts.

\textbf{Spatiotemporal tube (drifting):}
We mask a thin block that persists across consecutive time steps and drifts slowly: $\mathbf{M}(t_k,h,w)=1$ for $h\in[h_k,h_k{+}r_h)$, $w\in[w_k,w_k{+}r_w)$ with $t_k=t_0{+}k$, and bounded motion $|h_{k+1}{-}h_k|\le\delta_h$, $|w_{k+1}{-}w_k|\le\delta_w$. This enforces temporally coherent interpolation along physically plausible trajectories.

\textbf{Pilot-lattice (comb):}
To emulate sparse sounding, we reveal a periodic subset over $(t,w)$ shared across angles: $\mathbf{M}(t,h,w)=0$ if $t\equiv o_t\ (\mathrm{mod}\ S_t)$ and $w\equiv o_w\ (\mathrm{mod}\ S_w)$, and $\mathbf{M}(t,h,w)=1$ otherwise for all $h$. The model fills the missing entries from angular structure and temporal dynamics.

\textbf{Uniform random:}
We uniformly sample $K$ tokens to mask as a mode-agnostic regularizer.

\textbf{Curriculum learning:}
We gradually increase $\rho$ during training, moving from mild occlusion (learning coarse structure) to severe occlusion (requiring long-range spatiotemporal reasoning). Overall, these operators target common failure modes: local occlusions (rect), temporally coherent missing regions (tube), and structured sparse observations (comb).

\subsection{Loss Function}
\label{sec:pretraining_loss}

Masked tokens are reconstructed with a normalized objective that emphasizes salient energy as follows
\begin{align}
\mathcal{L}_{\mathrm{NMSE}}
\;=\;
\frac{1}{|\mathcal{M}|}
\sum_{i\in\mathcal{M}}
\frac{\|\mathbf{x}_i-\hat{\mathbf{x}}_i\|_2^2}{\|\mathbf{x}_i\|_2^2+\epsilon},
\end{align}
where $\mathcal{M}$ indexes the masked set. The normalization prevents low-energy regions from dominating the loss and sharpens fidelity on dominant paths, which is critical in the angle-delay domain where most tokens carry negligible energy.

\subsection{Data Generation (Dynamic Digital Twin)}
\label{subsec:dynamic_dt}

We construct a dynamic wireless digital twin by augmenting static DeepMIMO ray-tracing outputs~\cite{alkhateeb2019deepmimogenericdeeplearning} with mobility and Doppler evolution. At each UE grid point $\mathbf{p}$, DeepMIMO provides a set of multipath components; for each path $\ell$, it specifies the complex gain $\alpha_\ell(\mathbf{p})$, delay $\tau_\ell(\mathbf{p})$, and departure/arrival angles $\theta_\ell^{\mathrm{AoD}}(\mathbf{p})$ and $\phi_\ell^{\mathrm{AoA}}(\mathbf{p})$. We then convert these geometry-anchored snapshots into time-indexed wideband channel sequences ${\mathbf{H}[t]}_{t=1}^{T}$ along realistic user trajectories.

\textbf{1. Mobility-constrained trajectories:} We impose motion constraints by constructing a directed road graph $\mathcal{G}=(\mathcal{V},\mathcal{E})$ over DeepMIMO grid points, where each vertex $\mathbf{p}\in\mathcal{V}$ is a valid UE location and each directed edge $(\mathbf{p}\!\to\!\mathbf{q})\in\mathcal{E}$ connects physically adjacent points consistent with a Manhattan-style layout (lane directions and intersections). Each actor generates a discrete trajectory $\mathbf{p}[t]\in\mathcal{V}$ by walking on $\mathcal{G}$; vehicles follow lane directions and sample turns at intersections, while pedestrians follow a constrained random walk with bounded heading changes. A speed profile $v[t]$ induces timestamps through the kinematic relation $\Delta s[t]\approx v[t]\Delta t$, enabling variable-speed and stop-and-go behavior.

\textbf{2. Doppler-consistent evolution:} Let $\mathbf{v}[t]$ denote the actor velocity vector at time $t$. For each MPC $\ell$, we compute the Doppler shift from the radial velocity as $f_{d,\ell}[t]=\frac{f_c}{c}\,\mathbf{v}[t]^\top \hat{\mathbf{r}}_\ell[t]$, where $f_c$ is the carrier frequency, $c$ is the speed of light, and $\hat{\mathbf{r}}_\ell[t]$ is the unit propagation direction associated with the MPC (derived from AoA/AoD under the link convention). The path delays $\tau_\ell[t]$ are taken directly from the DeepMIMO ray-tracing outputs along the trajectory (and, for sub-grid positions, from the same trajectory-conditioned interpolation of DeepMIMO-provided delays), ensuring geometry-consistent delay evolution without explicit path-length recomputation.

\textbf{3. Wideband MIMO-OFDM synthesis:} For subcarrier $k$ with frequency $f_k$, the $\ell$-th MPC contributes $
\mathbf{H}_\ell[t,k]=
\alpha_\ell[t]\,
e^{-j2\pi f_k\tau_\ell[t]}\,
e^{j2\pi f_{d,\ell}[t]\,t\Delta t}\,
\mathbf{a}_{\rm rx}\!\big(\phi_\ell^{\rm AoA}[t]\big)\,
\mathbf{a}_{\rm tx}\!\big(\theta_\ell^{\rm AoD}[t]\big)^{\!H}$, where $\mathbf{a}_{\mathrm{tx}}(\cdot)$ and $\mathbf{a}_{\mathrm{rx}}(\cdot)$ are the transmit/receive array responses defined by the scenario. The full channel is then $\mathbf{H}[t,k]=\sum_{\ell=1}^{L(\mathbf{p}[t])}\mathbf{H}_\ell[t,k]$.

\textbf{Sub-grid channel interpolation:} High-rate sampling or slow motion can require displacements $\Delta s[t]=\|\mathbf{v}[t]\|\Delta t$ below the ray-traced grid spacing, making dense ray tracing infeasible at scale. We therefore ray-trace on a coarse lattice with spacing $d_{\mathrm{grid}}=10$\,cm and generate intermediate channels by interpolating MPC parameters between consecutive nodes on the coarse trajectory. Concretely, if the actor moves from coarse node $\mathbf{p}[n]$ to $\mathbf{p}[n{+}1]$, any refined position lies on the segment $\mathbf{p}^\star=\mathbf{p}[n]+\lambda(\mathbf{p}[n{+}1]-\mathbf{p}[n])$ for some $\lambda\in[0,1]$, with weights $w_0=1-\lambda$ and $w_1=\lambda$. We then interpolate $\tau_\ell(\mathbf{p}^\star)=w_0\tau_\ell(\mathbf{p}[n])+w_1\tau_\ell(\mathbf{p}[n{+}1])$, $\alpha_\ell(\mathbf{p}^\star)=w_0\alpha_\ell(\mathbf{p}[n])+w_1\alpha_\ell(\mathbf{p}[n{+}1])$, and $\boldsymbol{\vartheta}_\ell(\mathbf{p}^\star)=w_0\boldsymbol{\vartheta}_\ell(\mathbf{p}[n])+w_1\boldsymbol{\vartheta}_\ell(\mathbf{p}[n{+}1])$ for angular parameters $\boldsymbol{\vartheta}_\ell$, using wrap-aware interpolation for periodic quantities. Channels at $\mathbf{p}^\star$ are then synthesized by the same superposition model, enabling sub-grid motion resolution without dense ray tracing.

\textbf{Dataset scope:} We generate sequences across multiple cities (Cape Town, New Taipei, Houston, Tempe, Casablanca, Shibuya, Chiyoda, Melbourne, Seoul, Dongcheng, and Philadelphia), with 10{,}000 sequences per city. Each sequence contains $T=20$ snapshots sampled at $\Delta t=1$\,ms, with constant velocity uniformly sampled from $[0,30]$\,m/s and carrier frequency $f_c=3.5$\,GHz. Each snapshot is represented as a $(32,32)$ angle-delay matrix. We use an 80/20 per-city split for pretraining and validation.

\section{Downstream Task}

\begin{table*}[t]
\centering
\caption{Channel prediction NMSE (dB; lower is better) across training sample percentages for ray-traced and stochastic datasets}
\label{tab:channel_pred}
\resizebox{\textwidth}{!}{%
\begin{tabular}{l l *{6}{c} *{6}{c}}
\toprule
\multicolumn{1}{c}{\bf Velocity (m/s)} &
\multicolumn{1}{c}{\bf Model} &
\multicolumn{6}{c}{\bf Ray-Traced Fine-tuning Samples (\%)} &
\multicolumn{6}{c}{\bf Stochastic Fine-tuning Samples (\%)} \\
\cmidrule(lr){3-8} \cmidrule(lr){9-14}
& & \textbf{0} & \textbf{1} & \textbf{2} & \textbf{10} & \textbf{50} & \textbf{100} &
\textbf{0} & \textbf{1} & \textbf{2} & \textbf{10} & \textbf{50} & \textbf{100} \\
\midrule

\multirow{8}{*}{[0,10)} 
  & \textbf{LWM}          &   \textbf{-15.218}    &   \textbf{-15.232}    &   \textbf{-15.297}    &   \textbf{-15.356}    &   \textbf{-15.618}    &   \textbf{-17.080}    &   \underline{-6.777}    &   \underline{-7.551}    &   \underline{-8.506}    &   -9.904    &   \underline{-15.512}    &   \underline{-19.658}    \\
  & WiFo-Tiny             &  11.242 &   3.591 &   3.586 &   0.621 &   0.070 &   0.070 &   3.693 &   2.777 &   2.467 &   1.398 &   0.193 &  -1.982 \\
  & WiFo-Base             &   1.105 &   0.039 &   0.037 &  -0.015 &  -6.891 &  -6.917 &   0.034 &   0.011 &  -0.484 &  -1.694 &  -7.557 & -12.696 \\
  & WiFo-Tiny-RT          &   3.634 &   0.155 &   0.059 &  -0.014 &  -0.060 &  -3.918 &   3.154 &   2.225 &   1.646 &   0.862 &   0.073 &  -5.520 \\
  & WiFo-Base-RT          &   1.105 &   0.037 &   0.029 &  -0.119 &  -7.016 &  -7.365 &   0.034 &   0.010 &  -0.495 &  -3.109 &  -9.408 & -13.101 \\
  & LSTM-PT               &  -4.085 &  -4.087 &  -5.167 &  -7.108 &  -7.111 &  -7.199 &  -6.105 &  -6.129 &  -6.756 &  \underline{-10.283} & -12.829 & -14.390 \\
  & GRU-PT                &  -2.331 &  -2.848 &  -2.868 &  -4.840 &  -5.339 &  -5.476 &  -3.054 &  -3.062 &  -3.628 &  -8.432 & -13.136 & -14.466 \\
  & S\&H (baseline)       &   \underline{-10.665}    &   \underline{-10.665}    &   \underline{-10.665}    &   \underline{-10.665}    &   \underline{-10.665}    &   \underline{-10.665}    &   \textbf{-34.417}    &   \textbf{-34.417}    &   \textbf{-34.417}    &   \textbf{-34.417}    &   \textbf{-34.417}    &   \textbf{-34.417}    \\
\midrule

\multirow{8}{*}{[10,20)} 
  & \textbf{LWM}          &   \textbf{-12.664}    &   \textbf{-12.717}    &   \textbf{-12.782}    &   \textbf{-13.038}    &   \textbf{-13.226}    &   \textbf{-14.045}    &   \underline{-6.446}    &   \underline{-7.237}    &   \underline{-8.266}    &   \underline{-9.609}    &   \underline{-15.311}    &   \underline{-18.080}    \\
  & WiFo-Tiny             &  11.173 &   7.987 &   7.941 &   2.830 &   0.432 &   0.143 &   3.841 &   2.265 &   2.222 &   1.245 &   0.194 &  -1.293 \\
  & WiFo-Base             &   1.094 &   0.482 &   0.235 &   0.099 &   0.063 &  -4.148 &   0.062 &  -0.097 &  -0.282 &  -1.271 &  -8.223 & -12.402 \\
  & WiFo-Tiny-RT          &   3.583 &   0.711 &   0.382 &   0.137 &   0.130 &  -3.535 &   3.286 &   1.127 &   1.103 &   0.985 &  -0.210 &  -7.463 \\
  & WiFo-Base-RT          &   1.094 &   0.533 &   0.227 &   0.158 &   0.066 &  -5.058 &   0.062 &  -0.224 &  -0.404 &  -2.353 &  -8.867 & -13.785 \\
  & LSTM-PT               &  \underline{-4.085} &  \underline{-4.151} &  \underline{-5.482} &  \underline{-7.151} &  \underline{-7.301} &  \underline{-7.352} &  -6.035 &  -6.083 &  -6.738 &  -9.589 & -12.987 & -13.949 \\
  & GRU-PT                &  -2.331 &  -2.458 &  -2.621 &  -4.658 &  -4.481 &  -5.035 &  -2.827 &  -2.834 &  -3.324 &  -7.764 & -12.393 & -13.428 \\
  & S\&H (baseline)       &   -1.660    &   -1.660    &   -1.660    &   -1.660    &   -1.660    &   -1.660    &   \textbf{-25.582}    &   \textbf{-25.582}    &   \textbf{-25.582}    &   \textbf{-25.582}    &   \textbf{-25.582}    &   \textbf{-25.582}    \\
\midrule

\multirow{8}{*}{[20,30)} 
  & \textbf{LWM}          &   \textbf{-7.993}    &   \textbf{-8.088}    &   \textbf{-8.403}    &   \textbf{-8.620}    &   \textbf{-11.009}    &   \textbf{-11.545}    &   \underline{-6.148}    &   \underline{-6.930}    &   \underline{-7.955}    &   -9.191    &   \underline{-14.728}    &   \underline{-15.772}    \\
  & WiFo-Tiny             &  11.742 &   9.096 &   9.024 &   3.563 &   0.577 &   0.171 &   3.566 &   2.139 &   2.117 &   1.181 &   0.201 &  -0.750 \\
  & WiFo-Base             &   1.233 &   0.218 &   0.161 &   0.101 &   0.071 &  -0.595 &   0.066 &  -0.024 &  -0.336 &  -1.154 &  -8.055 & -12.367 \\
  & WiFo-Tiny-RT          &   3.932 &   0.542 &   0.506 &   0.403 &   0.111 &   0.079 &   3.040 &   0.657 &   0.162 &   0.133 &  -0.048 &  -0.055 \\
  & WiFo-Base-RT          &   1.233 &   0.177 &   0.173 &   0.238 &  -0.717 &  -4.723 &   0.066 &  -0.024 &  -0.032 &  -2.330 &  -8.927 & -14.133 \\
  & LSTM-PT               &  \underline{-4.085} &  \underline{-4.123} &  \underline{-5.651} &  \underline{-7.052} &  \underline{-7.069} &  \underline{-7.318} &  -6.057 &  -6.064 &  -6.651 &  \underline{-9.851} & -12.605 & -13.285 \\
  & GRU-PT                &  -2.331 &  -2.536 &  -2.598 &  -5.647 &  -5.805 &  -5.818 &  -2.937 &  -2.963 &  -3.496 &  -7.751 & -11.643 & -12.859 \\
  & S\&H (baseline)       &   1.834    &   1.834    &   1.834    &   1.834    &   1.834    &   1.834    &   \textbf{-21.343}    &   \textbf{-21.343}    &   \textbf{-21.343}    &   \textbf{-21.343}    &   \textbf{-21.343}    &   \textbf{-21.343}    \\
\bottomrule
\end{tabular}
}
\end{table*}

We evaluate \emph{channel prediction} as the main downstream task because it stress-tests whether a representation captures propagation-driven spatiotemporal dynamics under mobility (angle/delay drift, Doppler coupling, and blockage-induced birth/death). We perform one-step prediction using $T=10$ past frames across low, medium, and high velocity ranges: $[0,10)$, $[10,20)$, $[20,30)$ m/s. NMSE is reported after fine-tuning with $\{0,1,2,10,50,100\}\%$ of (i) ray-traced (RT) sequences from our dynamic digital-twin pipeline and (ii) stochastic 3GPP CDL data (1000 LoS/NLoS UMa samples). All methods are evaluated on a held-out RT test set from Parow, South Africa; WiFo-Tiny/Base, LSTM-PT, and GRU-PT are RT-pretrained for fair comparison.

\textbf{Propagation-aligned architecture:}
Table \ref{tab:channel_pred} shows LWM-Temporal achieves the best NMSE across all velocities and fine-tuning budgets, with the largest gains in the low-data regime. With $10\%$ RT fine-tuning, it reaches $-15.36$ dB (low velocity), surpassing competing models even at $100\%$ fine-tuning. With full RT fine-tuning, LWM-Temporal attains $-17.08$, $-14.05$, and $-11.55$ dB for low/medium/high velocities, while WiFo variants and recurrent baselines remain $3$--$5$ dB worse, indicating that architecture, not only pretraining scale, drives predictive performance.

\textbf{Data realism:}
RT fine-tuning transfers substantially better to the RT test distribution than stochastic CDL fine-tuning, especially at small budgets. For example, in the low-velocity bin, $50\%$ RT fine-tuning ($-15.62$ dB) outperforms $100\%$ stochastic fine-tuning for LWM-Temporal, with similar trends across models. Overall, robust forecasting under mobility requires \emph{both} geometry-consistent data and propagation-aligned inductive bias.

\section{Conclusion}

We introduced LWM-Temporal, a physics-informed foundation model for spatiotemporal wireless channels that combines sparse spatio-temporal attention (SSTA) with domain-aligned pretraining. By injecting geometric priors into attention and masking in the angle-delay-time domain, LWM-Temporal captures propagation-consistent dependencies at substantially lower cost. Experiments show improved NMSE across mobility levels and fine-tuning regimes, with stronger generalization to realistic dynamics than stochastic-model pretraining and prior recurrent/foundation baselines.

\section*{Acknowledgement}
This work was supported in part by the National Science Foundation under Grant No. 2426906.
    
\balance
\bibliographystyle{ieeetr}
% \bibliography{refs.bib} 

\end{document}